\documentclass[10pt,journal,compsoc]{IEEEtran}
\usepackage{amsmath}
\usepackage{graphicx}

\usepackage{xcolor}

\ifCLASSOPTIONcompsoc
  \usepackage[nocompress]{cite}
\else
  \usepackage{cite}
\fi
\ifCLASSINFOpdf
\else
\fi
\begin{document}
%
% paper title
% Titles are generally capitalized except for words such as a, an, and, as,
% at, but, by, for, in, nor, of, on, or, the, to and up, which are usually
% not capitalized unless they are the first or last word of the title.
% Linebreaks \\ can be used within to get better formatting as desired.
% Do not put math or special symbols in the title.
\title{A Unified Transformer-based Network for Multimodal Emotion Recognition}
%
%
% author names and IEEE memberships
% note positions of commas and nonbreaking spaces ( ~ ) LaTeX will not break
% a structure at a ~ so this keeps an author's name from being broken across
% two lines.
% use \thanks{} to gain access to the first footnote area
% a separate \thanks must be used for each paragraph as LaTeX2e's \thanks
% was not built to handle multiple paragraphs
%
%
%\IEEEcompsocitemizethanks is a special \thanks that produces the bulleted
% lists the Computer Society journals use for "first footnote" author
% affiliations. Use \IEEEcompsocthanksitem which works much like \item
% for each affiliation group. When not in compsoc mode,
% \IEEEcompsocitemizethanks becomes like \thanks and
% \IEEEcompsocthanksitem becomes a line break with idention. This
% facilitates dual compilation, although admittedly the differences in the
% desired content of \author between the different types of papers makes a
% one-size-fits-all approach a daunting prospect. For instance, compsoc 
% journal papers have the author affiliations above the "Manuscript
% received ..."  text while in non-compsoc journals this is reversed. Sigh.

\author{Kamran Ali and
        Charles E. Hughes,~\IEEEmembership{Member,~IEEE}
\IEEEcompsocitemizethanks{\IEEEcompsocthanksitem Kamran Ali and Charles E. Hughes are both with the Department
of Computer Science, University of Central Florida,
USA, Orlando, FL 32816.\protect\\
% note need leading \protect in front of \\ to get a newline within \thanks as
% \\ is fragile and will error, could use \hfil\break instead.
E-mail: kamran.ali@ucf.edu
%\IEEEcompsocthanksitem 
}% <-this % stops an unwanted space
\thanks{Manuscript received April 19, 2005; revised August 26, 2015.}}

% note the % following the last \IEEEmembership and also \thanks - 
% these prevent an unwanted space from occurring between the last author name
% and the end of the author line. i.e., if you had this:
% 
% \author{....lastname \thanks{...} \thanks{...} }
%                     ^------------^------------^----Do not want these spaces!
%
% a space would be appended to the last name and could cause every name on that
% line to be shifted left slightly. This is one of those "LaTeX things". For
% instance, "\textbf{A} \textbf{B}" will typeset as "A B" not "AB". To get
% "AB" then you have to do: "\textbf{A}\textbf{B}"
% \thanks is no different in this regard, so shield the last } of each \thanks
% that ends a line with a % and do not let a space in before the next \thanks.
% Spaces after \IEEEmembership other than the last one are OK (and needed) as
% you are supposed to have spaces between the names. For what it is worth,
% this is a minor point as most people would not even notice if the said evil
% space somehow managed to creep in.

% The paper headers
\markboth{Journal of \LaTeX\ Class Files,~Vol.~14, No.~8, August~2015}%
{Shell \MakeLowercase{\textit{et al.}}: A Unified Transformer-based Network for Multi-model Emotion Recognition}
% The only time the second header will appear is for the odd numbered pages
% after the title page when using the twoside option.
% 
% *** Note that you probably will NOT want to include the author's ***
% *** name in the headers of peer review papers.                   ***
% You can use \ifCLASSOPTIONpeerreview for conditional compilation here if
% you desire.

% The publisher's ID mark at the bottom of the page is less important with
% Computer Society journal papers as those publications place the marks
% outside of the main text columns and, therefore, unlike regular IEEE
% journals, the available text space is not reduced by their presence.
% If you want to put a publisher's ID mark on the page you can do it like
% this:
%\IEEEpubid{0000--0000/00\$00.00~\copyright~2015 IEEE}
% or like this to get the Computer Society new two part style.
%\IEEEpubid{\makebox[\columnwidth]{\hfill 0000--0000/00/\$00.00~\copyright~2015 IEEE}%
%\hspace{\columnsep}\makebox[\columnwidth]{Published by the IEEE Computer Society\hfill}}
% Remember, if you use this you must call \IEEEpubidadjcol in the second
% column for its text to clear the IEEEpubid mark (Computer Society jorunal
% papers don't need this extra clearance.)

% use for special paper notices
%\IEEEspecialpapernotice{(Invited Paper)}

% for Computer Society papers, we must declare the abstract and index terms
% PRIOR to the title within the \IEEEtitleabstractindextext IEEEtran
% command as these need to go into the title area created by \maketitle.
% As a general rule, do not put math, special symbols or citations
% in the abstract or keywords.
\IEEEtitleabstractindextext{%
\begin{abstract}
The development of transformer-based models has resulted in significant advances in addressing various vision and NLP-based research challenges. However, the progress made in transformer-based methods has not been effectively applied to biosensing research. This paper presents a novel Unified Biosensor-Vision Multi-modal Transformer-based (UBVMT) method to classify emotions in an arousal-valence space by combining a 2D representation of an ECG/PPG signal with the face information. To achieve this goal, we first investigate and compare the unimodal emotion recognition performance of three image-based representations of the ECG/PPG signal. We then present our UBVMT network which is trained to perform emotion recognition by combining the 2D image-based representation of the ECG/PPG signal and the facial expression features. Our unified transformer model consists of homogeneous transformer blocks that take as an input the 2D representation of the ECG/PPG signal and the corresponding face frame for emotion representation learning with minimal modality-specific design. Our UBVMT model is trained by reconstructing masked patches of video frames and 2D images of ECG/PPG signals, and contrastive modeling to align face and ECG/PPG data. Extensive experiments on the MAHNOB-HCI and DEAP datasets show that our Unified UBVMT-based model produces comparable results to the state-of-the-art techniques.  

\end{abstract}

% Note that keywords are not normally used for peerreview papers.
}

% make the title area
\maketitle

% To allow for easy dual compilation without having to reenter the
% abstract/keywords data, the \IEEEtitleabstractindextext text will
% not be used in maketitle, but will appear (i.e., to be "transported")
% here as \IEEEdisplaynontitleabstractindextext when the compsoc 
% or transmag modes are not selected <OR> if conference mode is selected 
% - because all conference papers position the abstract like regular
% papers do.
\IEEEdisplaynontitleabstractindextext
% \IEEEdisplaynontitleabstractindextext has no effect when using
% compsoc or transmag under a non-conference mode.

% For peer review papers, you can put extra information on the cover
% page as needed:
% \ifCLASSOPTIONpeerreview
% \begin{center} \bfseries EDICS Category: 3-BBND \end{center}
% \fi
%
% For peerreview papers, this IEEEtran command inserts a page break and
% creates the second title. It will be ignored for other modes.
\IEEEpeerreviewmaketitle

\IEEEraisesectionheading{\section{Introduction}\label{sec:introduction}}
% Computer Society journal (but not conference!) papers do something unusual
% with the very first section heading (almost always called "Introduction").
% They place it ABOVE the main text! IEEEtran.cls does not automatically do
% this for you, but you can achieve this effect with the provided
% \IEEEraisesectionheading{} command. Note the need to keep any \label that
% is to refer to the section immediately after \section in the above as
% \IEEEraisesectionheading puts \section within a raised box.

% The very first letter is a 2 line initial drop letter followed
% by the rest of the first word in caps (small caps for compsoc).
% 
% form to use if the first word consists of a single letter:
% \IEEEPARstart{A}{demo} file is ....
% 
% form to use if you need the single drop letter followed by
% normal text (unknown if ever used by the IEEE):
% \IEEEPARstart{A}{}demo file is ....
% 
% Some journals put the first two words in caps:
% \IEEEPARstart{T}{his demo} file is ....
% 
% Here we have the typical use of a "T" for an initial drop letter
% and "HIS" in caps to complete the first word.
\IEEEPARstart{O}{ver} the past few years, there has been an increasing interest in adopting a bio-sensing perspective for emotion analysis. Certainly, the popularity of bio-sensing research extends beyond affective computing. Various fields such as robotics \cite{1}, \cite{2}, health \cite{3}, \cite{4}, and virtual reality \cite{5} have also embraced bio-sensing as a valuable research tool. Bio-sensing systems, including those that measure electrocardiogram (ECG), Photoplethysmography (PPG), electroencephalogram (EEG), galvanic skin response (GSR), etc, have been available for decades; however, their size and complexity have limited their use to controlled laboratory settings and hospitals. The emergence of wearable biosensing systems has fueled the recent interest in utilizing bio-sensing systems for various applications by simplifying and speeding up data collection.

Several research studies have demonstrated the feasibility of recognizing human emotions through facial expressions captured in images and videos \cite{6}, \cite{7}, \cite{8}. Although facial expression recognition systems have shown remarkable success on  databases captured under controlled conditions, their performance significantly degrades \cite{9}, \cite{10} when they are applied to real-life situations. The system's unreliability in natural environments is primarily due to various factors, such as varying head pose, illumination conditions, and occlusion of different parts of the face. Additionally, recognizing emotions from facial expressions may not be entirely dependable since they can be easily concealed or manipulated. Furthermore, facial expressions can be influenced by social and cultural differences, as human expressiveness varies among individuals. In contrast, physiological signals offer a more accurate reflection of emotions and their subtle changes. Therefore, multimodal emotion recognition techniques that combine facial information with physiological data have the potential to compensate for the drawbacks of unimodal methods and achieve more precise recognition outcomes \cite{11}, \cite{12}.

\begin{figure}[t]
    \centering
    \includegraphics[width=9.0cm, height=6.0cm]{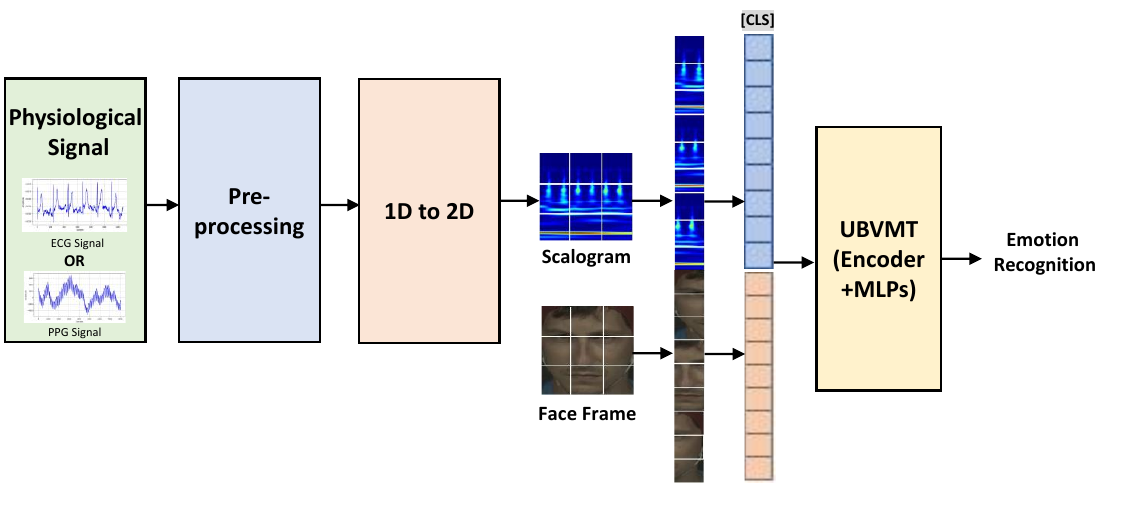}
    \vspace{-0.9cm}
    \caption{The framework of UBVMT-based multimodal emotion recognition}
    \label{fig:1.1}
    \vspace{-0.5cm}
\end{figure}

Many emotion recognition techniques have been proposed in the past years combining different modalities. For instance, in \cite{13} Shang et al. recognized emotions by fusing EEG, EOG, and EMG signals. Similarly, in \cite{14}, Miranda et al. combined EEG, ECG, and GSR modalities to perform emotion analysis. In \cite{15}, Koelstra et al. classified emotions along the valance and arousal axes by leveraging face information and EEG signals. Most of these multimodal emotion recognition methods combine different modalities with the EEG signal. However, several research studies show that there is a strong correlation between features extracted from the ECG signal and human emotions. For instance, in \cite{16}, Hany et al. extracted ECG features that represent the statistical distribution of dominant frequencies by applying spectrogram analysis to classify emotions. Similarly, in \cite{17}, time and frequency domain features are extracted from the ECG signal to perform emotion analysis. Building on those approaches, this paper presents a novel multimodal emotion recognition technique that fuses information from face video frames and ECG/PPG data. To the best of our knowledge, this is the first study that investigates the fusion of face information with the ECG/PPG signal to classify emotions.     

Many attempts have been made to create effective techniques for integrating multimodal information. Initially, methods involved simply combining high-level features from all modalities to predict outcomes (known as "early fusion") or adding up decisions made by each unimodal system with learnable weights (called "late fusion") to produce the final inference. While these methods outperformed unimodal techniques to some extent, the lack of inter-modality interactions during training limited the potential improvement. In recent times, Transformers \cite{18} have emerged as a highly successful approach for multimodel representation learning. Due to the advances in attention mechanisms and transformers, later studies have focused predominantly on using these techniques to develop more sophisticated multimodal fusion methods \cite{19}, \cite{20}, \cite{21}, \cite{22}, \cite{23}. However, most of these transformer-based multimodal emotion recognition techniques leverage video, audio, and text modalities to learn joint emotion representation. In this paper, we present a unified transformer-based model that employs ECG/PPG data along with face information to recognize emotions. The main framework of the proposed method is shown in Figure \ref{fig:1.1}. Based on our literature survey, no previous work has investigated the application of transformers in learning emotion representation by using face and ECG/PPG modalities.

Inspired by the compact architecture of the recently published Textless Vision-Language Transformer (TVLT) \cite{22}, we present a Unified Biosensor-Vision Multimodal Transformer (UBVMT) network to recognize emotions by fusing face and ECG/PPG data. We first investigated the effectiveness of various image-based representations of ECG/PPG signals for biosensor-based unimodal emotion recognition. Several previous methods have employed an image-based representation of an ECG/PPG signal for heart rate estimation \cite{24}, \cite{25}, \cite{26}, \cite{27}, \cite{28}. For instance, in \cite{24}, Song et al. obtained a spatiotemporal representation of a pulse signal in a time-delayed way by constructing a Toeplitz matrix. The Toeplitz matrix is then converted into an image that is fed to a CNN to estimate heart rate information. The simplicity of the Toeplitz representation allows it to preserve both the morphological and chronological details of the 1D pulse signal. As a result, the CNN can accurately extract the correct HR values from input feature images. 

In \cite{25}, average pooling is applied to various blocks within an ROI region. The spatiotemporal maps are then generated by arranging these temporal sequences into rows, which are then fed to a deep network to get HR values. In \cite{26}, the sample mean sequences of the R, G, and B channels from the ROI of videos are extracted, and Short-Time Fourier Transform (STFT) is applied to construct the 2D time-frequency representations of the sequences. In \cite{27}, a spatiotemporal map is first computed from the input video as a representation of the BVP signal, and the spatiotemporal images are then mapped to BVP signals using the generative adversarial learning technique. In \cite{28}, Khare et al. converted the time-domain EEG biosignals into a time-frequency representation by employing Smoothed Pseudo Wigner-Ville distribution (SPWVD). 

Apart from the image-based representation of ECG signals for heart rate estimation, some works \cite{29} and \cite{30} have extracted deep learning features from the scalogram of biosensor signals to recognize cognitive load and hypertension risk stratification, respectively. They converted the PPG signals to Scalograms by using Continuous Wavelet Transform. Scalograms were used instead of spectrograms as the image representation for PPG signals. This is because scalograms provide better highlighting of the low-frequency or rapidly changing frequency components of the signal as compared to spectrograms.

In this paper, we first investigate and compare the performance of three image representations of the ECG/PPG signal for emotion recognition. More specifically, we perform unimodal emotion recognition by converting the time domain ECG/PPG signal into 1. spatiotemporal maps \cite{24}, 2. time-frequency representation by employing Smoothed Pseudo Wigner-Ville distribution (SPWVD) \cite{28}, and 3. Scalograms \cite{29} \cite{30}. We show that the best emotion recognition performance is obtained by employing the scalogram representation of the ECG and PPG signals. Therefore, our Unified Biosensor-Vision Multimodal Transformer (UBVMT) network is trained and validated by using the scalogram representation of the ECG/PPG signal along with the face images. The effectiveness of the proposed UBVMT-based multimodal emotion recognition technique is validated on two datasets: the MAHNOB-HCI \cite{31} and DEAP \cite{32} datasets. Extensive experiments show that the proposed approach produces comparable results to the state-of-the-art emotion recognition techniques. Overall, the main contributions of this paper are as follows:  

\begin{itemize}

\item Comparison of the performance of three image-based representations, namely spatiotemporal maps, time-frequency representation, and scalograms of the ECG/PPG signal for emotion recognition, and evidence that scalograms better represent the emotion features contained by the ECG/PPG signals. 

\item A novel technique that employs transformer architecture for biosensor-based multimodal emotion recognition.

\item The Unified Biosensor-Vision Multimodal Transformer (UBVMT) network consisting of homogenous blocks that learn vision-and-biosensor emotion representation with minimal modality-specific design, thus, making it compact and applicable in real-time emotion analysis.

\item Experimental results showing that the proposed Unified Biosensor-Vision Multimodal Transformer-based (UBVMT) technique learns effective multimodal emotion representation by fusing face with ECG/PPG data. 

\item A baseline for emotion recognition research created by the fusion of face data with an ECG/PPG signal.

\end{itemize}

\section{Related Work}
\begin{figure}[t]
    \centering
    \includegraphics[width=\linewidth]{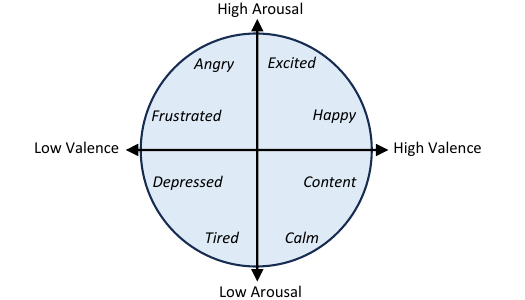}
    %\vspace{-0.7cm}
    \caption{Arousal valence emotion model}
    \label{fig:1}
    \vspace{-0.5cm}
\end{figure}

Psychologically, human emotions can be characterized based on two main frameworks: categorical or dimensional representations. In the categorical framework of emotions, emotions are classified into distinct labels such as joy, sadness, anger, happiness, fear, and so on. This approach considers emotions as discrete categories rather than continuous dimensions. In dimensional conceptualizations of emotions, a commonly used framework involves representing emotions within a two-dimensional space. In this space, valence is positioned along one axis, while arousal is positioned along the other \cite{33}. Valence, often positioned on the horizontal axis, signifies the extent of pleasantness or unpleasantness. On the other hand, arousal, typically placed on the vertical axis, represents the level of activation or energy associated with the emotion. Within a two-dimensional (2D) space, as shown in Figure \ref{fig:1}, emotions are portrayed by their valence and arousal level. While the categorical approach to emotions is conceptually straightforward, it faces certain challenges. It struggles to represent compound emotions that do not fit neatly into a single category, and it does not provide a means to quantify the degree or intensity of an emotional state. Therefore, in this paper, a novel multimodal emotion recognition technique is developed to classify emotions along the valence-arousal axis.

\subsection{Emotion Recognition with ECG/PPG Signals}
Previous research shows that there is a strong correlation between ECG/PPG signals and human emotion. In  \cite{34}, Yu et al. converted the rPPG signal of an input image into time-frequency domain spectrogram images to classify the short-term emotions. In  \cite{35}, Ismail et al. employed ECG and PPG signals to develop an emotion recognition system, and compared the performance of both signals for classifying emotions. Lee et al. in  \cite{36} investigated the ability of PPG signals to recognize emotions along the arousal-valence axis. Sepúlveda el al. in  \cite{37} performed emotion recognition from ECG signals using a wavelet scattering algorithm. Emotion features of the ECG signal are extracted at different time scales leveraging the wavelet scattering technique, and the extracted features are then fed into different classifiers to perform emotion analysis. Sarkar et al. in  \cite{38} exploited a self-supervised deep multi-task learning framework to develop an emotion recognition algorithm using the ECG signals. Mellouk et al. presented a rPPG signal-based emotion classification method using a deep learning architecture, which combines a one-dimensional convolution neural network (1DCNN) and a long short-term memory (LSTM)  \cite{39}. In  \cite{16}, Ferdinando et al. extracted emotion features from ECG signals using spectrogram analysis of intrinsic mode function after applying the bivariate empirical mode decomposition to ECG. Similarly, in  \cite{17}, Ferdinando et al. derived heart rate variability (HRV) features from ECG signals to categorize emotions in an arousal-valance space. In  \cite{29}, Gasparini et al. used a scalogram-based representation of a PPG signal to extract deep learning features for the recognition of cognitive load. Gasparini et al. showed that the deep learning features outperformed hand-crafted features to classify cognitive load,  especially by leveraging feature selection methods to avoid the curse of dimensionality. Similarly, Liang et al. in  \cite{30}, converted the PPG signals into scalogram-based representations using continuous wavelet transforms that are input to a deep learning method for the classification and evaluation of hypertension.

\subsection{Emotion Recognition with Facial Images}
Facial images and videos have been widely used by the research community to recognize emotions. Mostly, facial information is used to analyze emotions by using facial expression recognition techniques \cite{40}, \cite{41}, \cite{42}, \cite{43}, \cite{44}. However, in this paper, a transformer-based technique is presented that estimates the continuous dimensions related to emotions. Based on the dimensional approach, affective behavior can be characterized using various underlying continuous dimensions. These dimensions offer a more accurate representation of the emotions individuals experience in their daily lives. As noted in Figure \ref{fig:1}, two widely utilized dimensions are valence, which reflects the positivity or negativity of an emotional state, and arousal, which gauges the intensity of emotional activation \cite{45}, \cite{46}, \cite{47}.

To estimate valence and arousal from facial images, Dimitrios et al. \cite{48} used a Convolutional and Recurrent (CNN-RNN) deep neural architecture to extract emotion features for dimensional emotion recognition. Deng et al. \cite{49} used the ResNet-50 model for multi-task expression recognition and implemented the teacher-student architecture to enhance the training data without labels. This approach aims to address the imbalanced data distribution across different tasks. In \cite{50}, Kuhnke et al. utilized facial landmarks to align the image and eliminate conflicting data. They achieve this by leveraging the correlation among different representations to generate pseudo-labels. In \cite{51}, the features extracted from a 3D-CNN are inputted into the 3D VGG and 2D SENet-101 networks. Subsequently, Gated Recurrent Units (GRUs) are employed to enhance the model's ability to learn temporal features.

\subsection{Multi-modal Emotion Recognition}
There have been many studies focusing on the analysis of multimodal systems for recognizing human emotions. These studies have demonstrated that, by integrating information from multiple modalities, the performance of models for recognizing emotions can be enhanced \cite{52}. Therefore, in \cite{53}, Yin et al. fused information from ECG signal with Electroencephalogram (EEG), Electrooculography (EOG), Galvanic Skin Response (GSR), Electromyography (EMG), skin temperature, blood volume, and respiration to recognize emotions. Miranda et al. \cite{14} recorded EEG, GRS, and ECG signals using wearable sensors and combined EEG and GRS information with the ECG signal to categorize human emotions. Soleymani et al. \cite{31} employed Hidden Markov Models (HMMs) to classify emotions by integrating ECG features with EEG, GSR, respiration, and skin temperature data. In \cite{54}, Stamos et al. fused emotion features extracted from the ECG signal with the features obtained from the EEG signal to perform emotion analysis. Santamaria et al. \cite{55} integrated the modalities of ECG and GSR to recognize emotions by employing Deep Convolutional Network (DCNN). Elalamy \cite{56} performed emotion analysis using the spectrogram and recurrence plots (RP) of ECG, EDA, and Photoplethysmography (PPG) signals individually, and investigated the performance by combining this information for multi-modal emotion recognition. Most of the aforementioned techniques fuse the ECG signal with other modalities such as EEG, EMG, GSR, etc., for multimodal emotion analysis. But, unlike the EEG signal, which has been combined with the facial information as proposed in \cite{15}, \cite{57}, \cite{58}, the ECG signal has not been fused with facial features to perform emotion recognition. In this paper, a Unified Biosensor-Vision Multimodal Transformer-based (UBVMT) emotion recognition technique is presented that integrates ECG/PPG data with facial information.

\section{RESEARCH METHODS}

This section discusses various 1D-to-2D transformation methods to extract emotion features from 2D representation of ECG/PPG signals. We then present our Unified Biosensor-Vision Multimodal Transformer (UBVMT) architecture that leverages the 2D representation of ECG/PPG signals along with face information to recognize emotions.

\begin{figure}[t]
    \centering
    \includegraphics[width=9cm, height=4.5cm]{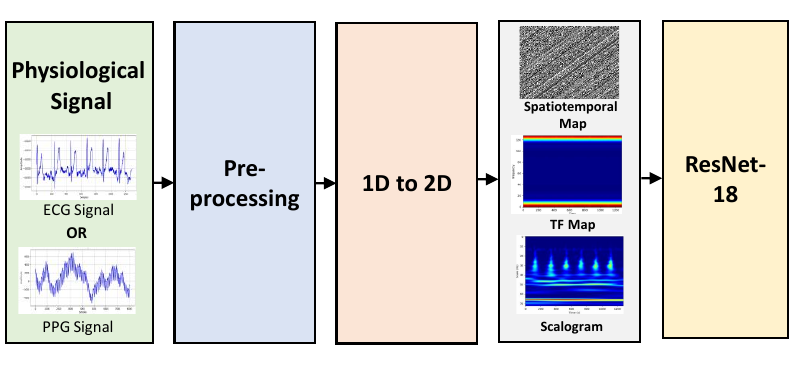}
    \vspace{-0.9cm}
    \caption{Unimodal emotion recognition}
    \label{fig:2}
    \vspace{-0.5cm}
\end{figure}

\subsection{The 2D Representation of ECG/PGG Signals}
In this paper, one of our goals is to find an effective method that can be used to transform the 1D time domain signal of ECG/PPG data into an image-based representation for emotion recognition. Therefore, in this section, we investigate the effectiveness of three different 1D-to-2D  transformation techniques to perform 2D image-based emotion analysis using ECG/PPG data. More specifically, we present a unimodal emotion recognition network using ResNet-18 \cite{62} by converting the time domain ECG and PPG signals into (1) spatiotemporal maps \cite{24}, (2) time-frequency representation by employing Smoothed Pseudo Wigner-Ville distribution (SPWVD) \cite{28}, and (3) Scalograms \cite{29}, \cite{30} as shown in Figure \ref{fig:2}.

\subsubsection{Spatiotemporal Representation of ECG/PPG signal}

In this section, a novel emotion recognition technique using a spatiotemporal representation of an ECG/PPG signal is presented. Rencheng et al. \cite{24} converted a 1D ECG and PPG signal into a 2D spatiotemporal map to estimate heart rate information. It has been reported that the performance degradation of the conventional PPG signals for heart rate estimation due to noise can be overcome by employing deep learning techniques on the spatiotemporal maps of the PPG signal \cite{24}. Therefore, in this section, we apply the technique proposed in \cite{24} to convert the 1D ECG/PPG signal into a 2D spatiotemporal map and perform emotion analysis using the spatiotemporal maps as input to the classifier. The spatiotemporal feature maps can retain both the morphological and chronological features of ECG/PPG signals. \newline  

%\subsubsection{Construction of ECG/PPG Spatiotemporal Map}

\textit{Construction of ECG/PPG Spatiotemporal Map:}
The spatiotemporal feature map of an ECG/PPG signal is constructed by creating a square Toeplitz matrix. Suppose the 1D input signal $S = (s_1, s_2, . . . , s_P )$ has P samples and P is an even number. The first row of the matrix consists of $s_1$ to $s_{P/2}$ samples. Similarly, the second row contains samples from the second point, i.e., $s_2$ to the $(P /2 + 1)th$ sample, and so on. Therefore, a square Toeplitz matrix T with a size equal to $P/2$ is constructed as: 
\vspace{-0.5cm}
\begin{center}
    % Your equation with matrix in large brackets
    \[
    T = \begin{bmatrix}
        s_1 & s_2 & ... & s_{P/2} \\
        s_2 & s_3 & ... & s_{P/2+1} \\
        .   & .   & .   & . \\
        .   & .   & .   & . \\
        .   & .   & .   & . \\
    s_{P/2} & s_{P/2+1} & ... & s_{P-1} \\
    \end{bmatrix}
    \]
\end{center}

A gray image is constructed by converting the matrix T into an image representation. The obtained gray image has a clear structure because the input signal is quasiperiodic. The second row of Figure \ref{fig:3} shows the spatiotemporal image representation of 1D signals. As it can be seen in Figure \ref{fig:3} that the vertical patterns preserve the period information of the 1D signal. This suggests that the 2D Toeplitz representation can effectively capture and portray the periodic nature of a 1D signal. The morphological and chronological information of the 1D signal is preserved in this simple 2D spatiotemporal representation. Therefore, emotion features can be extracted from these 2D maps of the ECG/PPG signals. The spatiotemporal image representations of ECG/PPG signals were then adjusted to the size of 224 × 224 × 3, and fed to a pre-trained ResNet-18 \cite{62} to perform emotion analysis.

\subsubsection{Time-frequency Representation of ECG/PPG Signal }

To leverage the effectiveness of Convolutional Neural Networks (CNNs), Khare et al. \cite{28} converted 1D EEG signals into 2D maps using a Time–Frequency (TF) representation for emotion recognition. More specifically, filtered EEG time-domain signal is transformed into time, frequency, and amplitude representation by employing Smoothed Pseudo-Wigner–Ville Distribution (SPWVD). The transformation of time-domain signals into Time-Frequency (TF) representation enables the preservation of signal information in the spectral domain. TF map is a representation that combines time, frequency, and amplitude in a spatial format simultaneously. Therefore, a time-frequency representation of the ECG/PPG signal can also be obtained by using other time-frequency analysis techniques such as Short-Time Fourier Transform (STFT), Wigner–Ville distribution, and so on. The time-frequency representation obtained by using STFT is known as a spectrogram. To perform Short-Time Fourier Transform (STFT), certain parameters need to be defined, such as the window size, shape, and sampling frequency. It is crucial to maintain a consistent window length across the entire signal. However, the resulting spectrogram from STFT often exhibits limited resolution due to the trade-off between time and frequency localization. In contrast, the utilization of the Wigner-Ville distribution for obtaining the time-frequency representation results in the presence of cross-terms and attenuation for low frequencies. As reported in \cite{59}, SPWVD offers enhanced time-frequency resolution, effectively resolving the limitations associated with STFT. The challenges associated with the Wigner-Ville distribution are tackled in SPWVD by incorporating a cross-term reducing window in the frequency domain.

\begin{figure}[t]
\centering
\includegraphics[width=8.5cm, height=8.0cm]{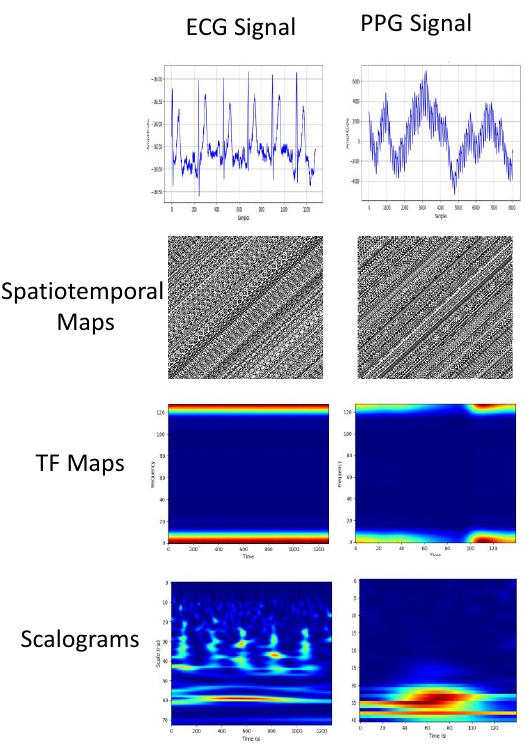}
\vspace{-0.3cm}
\caption{2D representations of ECG/PPG signal}
\label{fig:3}
\vspace{-0.5cm}
\end{figure}

In this paper, one of our goals is to find an effective method that can be used to transform the 1D time domain signal of ECG/PPG into an image-based representation. Therefore, to address the aforementioned constraints of STFT and the Wigner-Ville distribution, in this section, the technique of Smoothed Pseudo-Wigner–Ville Distribution (SPWVD) is employed to convert time-domain ECG/PPG signals into a time-frequency representation. The performance of a SPWVD-based TF representation of ECG/PPG signals is then investigated for emotion recognition. \newline  

%\subsubsection{Construction of ECG/PPG TF Map}
\textit{Construction of ECG/PPG TF Map:}
SPWVD provides a straightforward depiction of the localization of signal energy in both time and frequency domains. The choice of the length and type of the cross-term reducing window in both the time and frequency domains can be made independently. Due to the independent selection of the length and type of the cross-term reducing window, SPWVD exhibits favorable time-frequency cluster characteristics. The representation of SPWVD in mathematical terms can be expressed as \cite{59}, \cite{60}.

\begin{equation}
    SPWVD (t, f) = \int_{t_1} \int_{f_1}  u(t-t_1) v(f-f_1) W(t_1 f_1), dt_1 df_1
\end{equation}

\begin{equation}
    W(t, f) = \int_{-\infty}^{\infty} x(t+\frac{\tau}{2}) x^*(t-\frac{\tau}{2}) e^{-j 2 \pi f \tau} d \tau
\end{equation}

\noindent where $u(t)$ represents the smoothing window in the time domain, $v(f)$ denotes the smoothing window in the frequency domain, $\tau$ corresponds to a lag, and $x(t)$ is the input signal. Controlling the smoothing scales in both the time and frequency domains is straightforward. The length of the windows for $u(t)$ and $v(f)$ can be chosen independently. The TF representation of time-domain ECG/PGG signals obtained by employing the SPWVD technique is shown in the third row of Figure \ref{fig:3}.

To mitigate cross-terms in both time and frequency domains, the Kaiser window is employed. However, selecting a window size that is too small may lead to diminished resolution, while overly large windows can significantly increase the image size. Consequently, a medium-sized window with a length of 31, as suggested in \cite{28}, is chosen. For efficient computation, the window size is maintained at $2^n$-1, where n represents the number of bits. The TF representations of time-domain ECG/PGG signals are then adjusted to the size of 224 × 224 × 3, and fed to a pre-trained ResNet-18 \cite{62} to perform emotion recognition.

\subsubsection{Scalogram of ECG/PPG Signal }

Gasparini et al. \cite{29} converted the monodimensional photoplethysmography (PPG) data into a bidimensional representation, and applied a pre-trained CNN to classify the different levels of the subject’s hypertension. More specifically, the 1D PPG signals were converted into scalograms by leveraging Continuous Wavelet Transform \cite{61}. Gasparini et al. opted to utilize the scalogram instead of the spectrogram as the image representation for PPG signals. The reason for this choice is that scalograms offer enhanced highlighting of low-frequency or rapidly changing frequency components in the signal, compared to spectrograms. Similarly, in \cite{30}, Liang et al. converted 1D PPG signals into 2D scalograms by using Continuous Wavelet Transform to classify hypertension. In this paper, we investigate the performance of scalogram representation of ECG/PPG signals for deep learning-based emotion recognition. \newline 

%\subsubsection{Construction of ECG/PPG scalograms}
\textit{Construction of ECG/PPG scalograms:}
Continuous Wavelet Transform (CWT) has been used for decades as a valuable technique for analyzing both time and frequency information. In this paper, each segment of the ECG/PPG signal was transformed into a time-frequency representation, known as a scalogram, using the continuous wavelet transform method. A scalogram represents the absolute values of the continuous wavelet transform coefficients of a signal, displayed as a graph of time and frequency. In contrast to a spectrogram, a scalogram provides improved capability in identifying the low-frequency or rapidly changing frequency components of the signal. To convert the PPG signal into a scalogram, we obtained the absolute values of the wavelet coefficients for each signal segment using the analytic morse (3,60) wavelet. The scalogram representation of ECG/PGG signals obtained by employing the CWT technique is shown in the fourth row of Figure \ref{fig:3}. The scalograms were then adjusted to the size of 224 × 224 × 3, and fed to a pre-trained ResNet-18 \cite{62} to perform emotion recognition. 

\begin{figure*}[t]
    \centering
    \includegraphics[width=13.0cm, height=8.0cm]{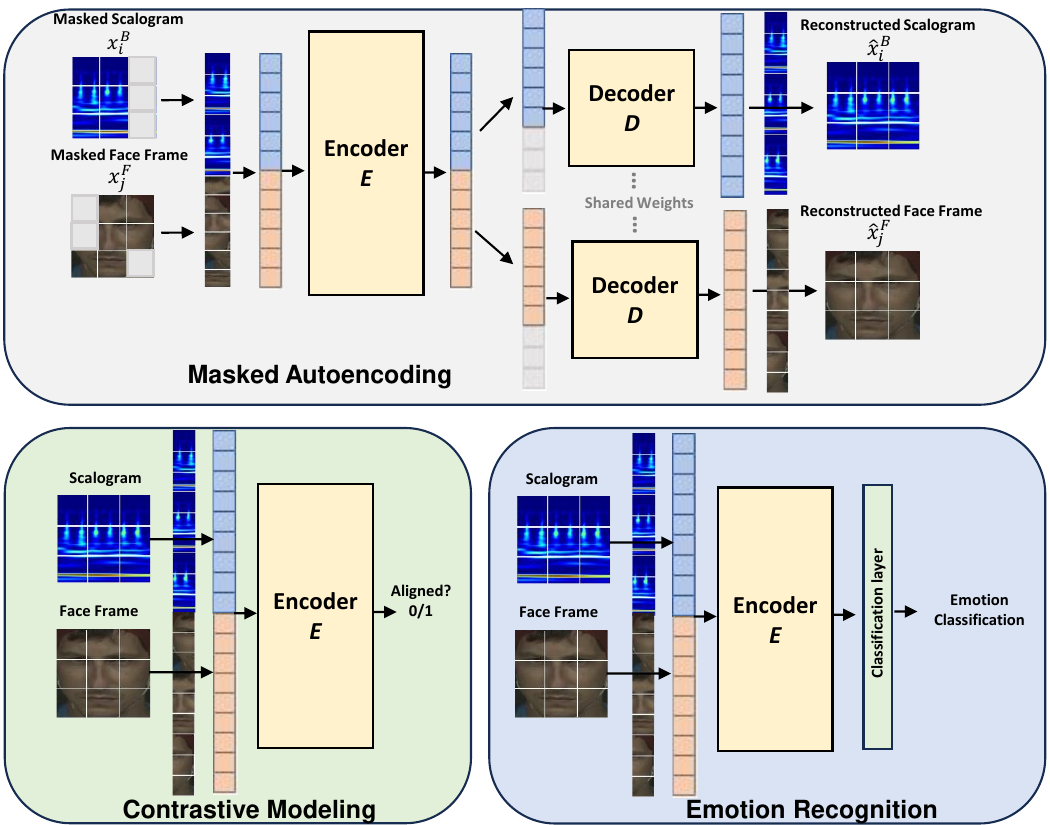}
    \vspace{-0.3cm}
    \caption{Multimodal emotion recognition}
    \label{fig:4}
    \vspace{-0.5cm}
\end{figure*}

\subsection{The Unified Transformer-based Multi-modal Network}
In this section, we discuss our Unified Biosensor-Vision Multimodal Transformer (UBVMT) network where homogeneous transformer blocks take 2D representations of ECG/PPG signals and raw visual inputs for multi-modal emotion representation learning with minimal modality-specific design. UBVMT is trained by employing masked autoencoding \cite{63} and contrastive modeling \cite{22} to learn effective emotion representation. The overall architecture of our Unified Biosensor-Vision Multi-modal Transformer (UBVMT) network is shown in Figure \ref{fig:4}. The objective of masked autoencoding is to reconstruct masked patches of 2D representations of ECG/PPG signals and face video frames, while contrastive modeling is applied to align ECG/PPG and face information. We argue that, due to the unified architecture of UBVMT, the computational redundancy and complexity of our method is reduced as compared to conventional transformer-based multi-modal emotion recognition techniques.

The input to the UBVMT is the integration of face patch embedding, ECG/PPG patch embedding, and modality embedding. To obtain face embeddings, the face image of
$224 \times  224 \times 3$ pixels is divided into a list of $16 \times 16$ sized patches. The pixel values of each patch are normalized, and a linear projection layer is used to convert face patches into a 768-dimensional patch embedding. For an input face image of size $224\times 224 \times 3$, the size of the resultant face embedding is $14 \times 14$. The spatial embedding involves incorporating spatial information for each input patch by adding a distinct trainable vector to the height and width axes of the $14 \times 14$ embeddings. For ECG/PPG embedding, the 2D representation of an ECG/PPG signal is divided into patches, and a liner projection layer is applied on each patch to obtain a 768-dimensional patch embedding. Similar to the face modality, a patch size of $16 \times 16 $ is used, and trainable temporal and frequency embeddings are utilized to denote the temporal and frequency information of the patches.

\subsubsection{The architecture of UBVMT}

Similar to the transformer model proposed in \cite{18}, UBVMT has an encoder E which has 12 layers (hidden size 768), and a decoder D with 8 layers (hidden size 512). After pretraining, we only employ the encoder $E$ part of the transformer and finetune it for emotion recognition. 
\subsubsection{Pretraining UBVMT}

UBVMT is pre-trained by employing masked autoencoding \cite{63} and contrastive modeling \cite{22} to learn effective emotion representation. The pretraining objective function $\mathcal{L}$ is the weighted sum of masked autoencoding loss $\mathcal{L_M}$ and the contrastive modeling loss $\mathcal{L_C}$:

\begin{equation}
    \mathcal{L} = \lambda_M \mathcal{L_M} + \lambda_C \mathcal{L_C}
\label{eq:pretrain}
\end{equation}

\noindent where $\lambda_M = 0.4$ and $\lambda_C = 1$.  \\

\textit{Masked Autoencoding Objective Function:}
The main objective of masked autoencoding is to learn effective unimodal representations in biosensor-and-vision settings. This objective involves masking random patches of ECG/PPG images and the face video frames, allowing us to reconstruct missing inputs effectively. Specifically, we implement a random dropout mechanism on a portion of ECG/PPG embedding $x^B$ and face embedding $x^F$. Subsequently, we input the remaining patch embeddings to the encoder $E$. To generate the input to the decoder $D$, we add the dropped embeddings as trainable vectors labeled as [MASK] and position them in the same locations as the original inputs (indicated by gray boxes in Figure \ref{fig:4}. Additionally, the corresponding positional, and frequency embeddings, which are separately parametrized, are incorporated into the decoder input. The objective function of masked autoencoding is a mean squared error between the reconstructed and original ECG/PPG images and face video frames:

\begin{equation}
\mathcal{L_M} = \frac {1} {N_m^B}  \sum_{i\in \text{masked}} \|x_i^B - \hat{x}_i^B\|_2^2 + 
\frac {1} {N_m^F}  \sum_{j\in \text{masked}} \|x_j^F - \hat{x}_j^F\|_2^2
\end{equation}

\noindent where $N_m^B$ and $N_m^F$ are the number of masked patches for ECG/PPG images and face video frames, respectively. Loss $\mathcal{L_M}$ is computed only on masked patches. Note that the ECG/PPG and face output of the encoder are fed to the decoder separately.  

\textit{Contrastive Modeling Objective Function:}
The main objective of contrastive modeling is to perform ECG/PPG-vision matching and learn an effective cross-modal representation, as shown in Figure \ref{fig:4}. For every face image, a positive vision-ECG/PPG pair $(x^{F+}; x^B)$ is generated. Additionally, we form half of the vision-ECG/PPG pairs within a batch as mismatched (negative) pairs $(x^{F-}; x^B)$ by substituting the face images $x^{F+}$ with randomly selected face images $x^{F-} $ from the training dataset. 

Similar to previous multimodal transformers \cite{22}, \cite{64}, \cite{65}, \cite{66}, \cite{67}, we incorporate a linear layer with sigmoid activation as the classification head. This is applied to the encoder output of the first [CLS] token, resulting in the matching probability $p$ by employing binary cross-entropy loss:

\begin{equation}
    \mathcal{L_C} = - y \log p
\end{equation}

\noindent where the value of y is set to 1 when the input vision-ECG/PPG pair $(x^{F}; x^B)$ is positive (match), and 0 otherwise. Throughout the training process, $\mathcal{L_M}$ and $\mathcal{L_C}$ are computed using separate forward passes.

\section{EXPERIMENTAL DETAILS}
\subsection{Database Description}
The effectiveness of transformer models to learn multimodal representation is improved by pretraining them on large datasets \cite{65}, \cite{66}, \cite{67}. As such, UBVMT is pre-trained on the large,  comprehensive CMU-MOSEI \cite{68} dataset. Similar to \cite{24}, \cite{25}, \cite{69}, rPPG signals are extracted from the CMU-MOSEI video clips using the MTTS-CAN \cite{70} method. After pertaining, MAHNOB-HCI \cite{31} and DEAP \cite{32} datasets are used for multimodal emotion analysis.

The \textbf{CMU-MOSEI} dataset comprises 23,454 movie review clips, totaling over 65.9 hours of YouTube video content. It features contributions from 1000 speakers and encompasses 250 distinct topics. Videos having non-human faces and faces with more than $80^\circ$ head rotations from the frontal position are discarded from the dataset. From each video, multiple video clips of 1.1 secs are extracted, with the number of clips being dependent on the length of each video. For our case, 90,037 video clips are extracted from the entire CMU-MOSEI dataset. These video clips are then fed to the MTTS-CAN \cite{70}, and only the pulse signals from the MTTS-CAN are used to obtain the rPPG signals of these video clips.

The \textbf{MAHNOB-HCI} \cite{31} (Multimodal Human-Computer Interaction) dataset is a widely used publicly available dataset designed for research in affective computing and multimodal emotion recognition. The dataset includes 527 facial video recordings of 27 participants engaged in various tasks and interactions, while their physiological signals such as 32-channel electroencephalogram (EEG), 3-channel electrocardiogram (ECG), 1-channel galvanic skin response (GSR), and facial expressions are captured. The ECG signals were sampled at a rate of 256 Hz. The ECG data is precisely synchronized and aligned with the face video recordings. Similar to \cite{71}, only the EXG2 signal from the 3 ECG channel system is extracted for emotion analysis.

The Database for Emotion Analysis using Physiological Signals \textbf{(DEAP)} \cite{32} is a widely used and publicly available database designed for research in emotion analysis and affective computing. The DEAP database contains data from 32 participants, aged between 19 and 37 ($50\%$ female), who were recorded watching 40 one-minute music videos. Each participant was asked to evaluate each video by assigning values from 1 to 9 for arousal, valence, dominance, like/dislike, and familiarity. Face video was recorded for 22 out of the 32 participants, and the proposed UBVMT method is evaluated using this particular group of subjects. For each dataset, we conduct a subject-independent 10-fold cross-validation evaluation.

\subsection{Pre-processing Steps}

The MAHNOB-HCI emotion elicitation data contains unstimulated baseline and stimulated response ECG signals. Furthermore, the database includes a synchronization signal that facilitates the separation of the two. Our experiments exclusively utilized the ECG signals recorded during the stimulated phase. To remove motion artifacts, the original signal is subtracted by the smoothing signal \cite{31}. Subsequently, similar to \cite{16}, \cite{17} a notch filter was applied at $60$ Hz to eliminate power line interference. Baseline drift was mitigated by implementing a highpass filter at $0.4$ Hz. Additionally, other noises were eliminated using a low-pass filter set at $200$ Hz. Due to the variability in the length of the ECG signals, they are partitioned into segments of 5 seconds each.

The PPG and rPPG signals from the DEAP and CMU-MOSEI datasets are preprocessed and segmented into segments of pulses corresponding to 1.1 secs following \cite{36}. PPG signal can be affected by disturbances, such as movement at the sensor attachment site. Consequently, PPG signals in the DEAP dataset contain movement noise. Hence, before segmenting PPG signals into single pulses of 1.1 secs, a high-order polynomial (an order of 50 polynomials) is fitted to the PPG signal. Subsequently, the fitted curve is subtracted from the original PPG signal, effectively eliminating any movement noise. To partition the PPG signal into single pulses of 1.1 secs, we first find the maximum or peaks of the signal and then use the peak value as the center of the segmenting window. The next pulse is segmented by moving the center of the segmenting window to the next peak, and so on. The biosignal data vary from person to person; as such, the PPG/rPPG signals are normalized to mitigate variations in PPG signal size among individuals. We must exercise caution to retain the personal characteristics of the signal, as they vary depending on the emotions. The PPG signals are normalized by employing personal maximum and minimum \cite{36}:

\begin{equation}
\bar{z_i} = \frac {(z_i - min_{person})} {(max_{person} - min_{person})}  \times \alpha
\end{equation}

\noindent where $z_i$ is the PPG signal, $\bar{z_i}$ is the normalized signal, and $\alpha$ is set to $1000$ to normalize $z_i$ between $0$ to $1000$. After extracting the signal segments and their corresponding video frames, we balance the dataset by discarding some segments that contain neutral facial expressions. We employ the off-the-shelf TER-GAN \cite{6} FER model trained on the in-the-wild AffectNet \cite{10} dataset to detect segments containing non-neutral facial expressions.
\subsection{Unimodal Emotion Recognition Details}
\label{sec:Unimodal}

This section presents the experimental details of the unimodal emotion recognition experiments that were performed to investigate and compare the performance of the three 2D representations of ECG/PPG signals. These experiments were conducted by inputting only the 2D image-based representations of the ECG signal (in the case of MAHNOB-HCI dataset), and the PPG signal (in the case of DEAP dataset) to a pre-trained ResNet-18 \cite{62}. ResNet-18 was employed to classify emotions in an arousal-valence space using both the MAHNOB-HCI and DEAP datasets. In the case of the MAHNOB-HCI dataset, the arousal dimension consisted of three categories: 'calm', 'medium', and 'activated', while the valence dimension included the categories 'unpleasant', 'neutral', and 'pleasant', similar to the class distribution presented in \cite{17}. The arousal and valence classes in the DEAP dataset are annotated on a 1-9 scale. Therefore, following \cite{36}, the arousal and valence are split into two binary classes based on a threshold of 5, indicating high and low arousal, and high and low valence, respectively.

ResNet-18 is finetuned by using Adam \cite{72} optimization algorithm with a learning rate of 0.001, batch size of 128, and cross-entropy loss function. The ResNet-18 is trained for 50 epochs, and 10-fold cross-validation is employed for the evaluation of the 2D ECG/PPG representation-based unimodal emotion recognition technique. The best emotion recognition performance is obtained by converting 1D ECG and PPG signals into the 2D scalogram representation. Further details and comparison of the performance of the three 2D representation techniques are discussed in section \ref{sec:unimodal}.  

\subsection{Multimodal Emotion Recognition Details}
Multimodal emotion recognition is performed by applying our novel Unified Biosensor-Vision Multi-modal Transformer (UBVMT) network for emotion recognition in an arousal-valence space using the facial expression information and 2D representation of ECG/PPG data from the MAHNOB-HCI and DEAP datasets, respectively. As far as we are aware, we are the first ones to model the co-relation between the face and ECG/PPG data for emotion recognition employing a transformer network. For the extraction of the face information, since the ECG and PPG signals are synchronized with the facial videos of the participants in both datasets, we partitioned the videos into clips of 5 secs for the MAHNOB-HCI dataset and 1.1 secs for the DEAP database. Subsequently, following \cite{7}, we extract the first frame from each video clip, pass it to MTCNN \cite{77} to extract only the face region as a facial expression representation, and input it to the UBVMT network along with the 2D representation of the corresponding ECG/PPG signal. As mentioned in section \ref{sec:Unimodal}, the best unimodal emotion recognition performance is obtained by transforming the 1D ECG and PPG signals into a 2D scalogram representation. Therefore, multimodal emotion analysis is performed by combining the facial expression images with the scalograms of ECG and PPG signals using our proposed UBVMT network. 

UBVMT is trained using the Adam \cite{72} optimizer, with a batch size of 4, learning rate of 1e-4, and using a cosine schedule \cite{73} with a decay rate set at 0.001. To formulate the pretraining objective in equation \ref{eq:pretrain}, the values of $\lambda_M$ and $\lambda_C$ are set to 0.4 and 1, respectively. Following MAE \cite{63} and TVLT \cite{22}, a random masking strategy is applied, where $75\%$ of the face and ECG/PPG patches are randomly masked. After pretraining, the encoder of UBVMT is detached, and a two-layer MLP is added on top of the encoder representation for emotion analysis. The encoder plus the MLP layers are fine-tuned using the Adam \cite{72} optimizer, the learning rate of 1e-4, and a decay rate of 0.001. All the training and validation processes are carried out using dual NVIDIA Tesla V100 GPUs.

\begin{table}
\caption{MAHNOB-HCI dataset: Performance comparison of 2D representations of ECG signal for unimodal emotion recognition.}
\vspace{-0.5cm}
\label{table:1}
\begin{center}
\begin{tabular}{|l| c| c|} 
 %\hline
 %\multicolumn{4}{|c|}{Country List} \\
 \hline
 Method&Valance&Arousal\\
 \hline
Spatio-temporal maps\cite{24}   & 37.62    &41.04\\
SPWVD\cite{28}&   38.39  & 42.75\\
 Scalogram\cite{29, 30}    &\textbf{42.91} & \textbf{49.14}\\
 \hline
\end{tabular}
\end{center}
\end{table}
%\vspace{-6.5cm}

\begin{table}
\caption{DEAP dataset: Performance comparison of 2D representations of PPG signal for unimodal emotion recognition.}
\vspace{-0.5cm}
\label{table:2}
\begin{center}
\begin{tabular}{|l| c| c|} 
 %\hline
 %\multicolumn{4}{|c|}{Country List} \\
 \hline
 Method&Valance&Arousal\\
 \hline
Spatio-temporal maps\cite{24}   & 60.26    &62.71\\
SPWVD\cite{28}&   64.03  & 65.19\\
 Scalogram\cite{29, 30}    &\textbf{76.51} & \textbf{77.02}\\
 \hline
\end{tabular}
\end{center}
\end{table}

\section{Results and Analysis}

In this section, the experimental results of both the unimodal emotion recognition and the multimodal emotion analysis using the proposed UBVMT method are discussed in detail.
\subsection{Results and Analysis of Unimodal Method}\label{sec:unimodal}
The unimodal emotion recognition performance of the three 2D representations of ECG/PPG signals for MAHNOB-HCI and DEAP datasets are presented in Table \ref{table:1} and Table \ref{table:2} as an average of the 10-fold cross-validation. As it can be seen, the scalogram representation of both ECG and PPG signals outperforms the emotion recognition performance of the spatiotemporal maps and the SPWVD-based 2D representation. The superior performance of scalogram representation is because it is more effective at identifying the low-frequency or rapidly-changing frequency components of the signal, as both ECG and PPG are low-frequency signals. Similarly, the emotion recognition accuracy of SPWVD is higher than the accuracy of spatio-temporal maps because SPWVD offers better time-frequency resolution, and thus captures emotion features more effectively than the spatio-temporal maps.

The performance of our unimodal emotion recognition method is also compared with the state-of-the-art unimodal emotion recognition techniques on both the MAHNOB-HCI and DEAP datasets, as presented in Table \ref{table:3}. For the MAHNOB-HCI dataset, we compare our result with the ECG-based emotion analysis technique proposed by Ferdinando et al. \cite{17} and rPPG-based method of Yu et al. \cite{76}. In \cite{17}, Ferdinando et al. extracted heart rate variability (HRV) features from ECG signals to classify emotions. As it can be seen in Table \ref{table:3}, our scalogram-based method outperforms the emotion recognition technique proposed in \cite{17}. In \cite{76}, Yu et al. extracted ten-dimensional HRV features from rPPG signals of MAHNOB-HCI video clips, and fed them to a support vector machine for emotion classification. Table \ref{table:3} shows that they achieve a higher recognition accuracy in the case of valence, while our scalogram-based method outperforms their method in arousal categorization. Similarly, in Table \ref{table:3}, the performance of our PPG-based emotion classification method is compared with the state-of-the-art emotion analysis techniques using the DEAP database. For the comparison of the techniques involving only the PPG signal, our 2D scalogram-based method is compared with the PPG-based technique proposed by Lee et al. \cite{36} and various 2D representations of PPG signal employed by  Elalamy et al. \cite{56}. Table \ref{table:3} shows that our scalogram-based PPG technique is more effective than the one-dimensional convolutional neural network-based (1D CNN) emotion analysis method proposed by Lee et al. \cite{36}. Similarly, our technique produces higher recognition accuracy than the 2D spectrogram-based method proposed by Elalamy et al. \cite{56}. Also, our 2D scalogram representation outperforms the emotion recognition performance of the Recurrence PLot-based (RP) 2D representation technique used in \cite{56}. Most of the physiological-based emotion analysis methods employ EEG signals for emotion recognition more than any other physiological signal due to its capability to capture emotion features more effectively. Therefore, in Table \ref{table:3}, we also compare our unimodal method with the state-of-the-art unimodal emotion analysis techniques using EEG signals. Table \ref{table:3} shows that our scalogram-based PPG representation produces higher recognition accuracy as compared to the EEG-based emotion recognition techniques of  Patras et al. \cite{32}, Chung et al. \cite{74} and Campos et al. \cite{75}. These results imply that the scalogram-based 2D representation of the PPG signal can be used as an alternative to the EEG signal. Obtaining and utilizing EEG signals, the most prevalent bio-signal in emotion recognition, can be inconvenient due to the high cost of EEG measurement devices and the cumbersome measurement process, especially for participants from neurodiverse populations such as kids with ASD. 

\begin{table}
\caption{Performance comparison of our unimodal emotion recognition method with the state-of-the-art techniques.}
%\vspace{-0.5cm}
\label{table:3}
\begin{center}
%\begin{tabular}{l p{4.5cm} c c} 
\begin{tabular}{l p{2.5cm} c c} 

 %\hline
 %\multicolumn{4}{|c|}{Country List} \\
 \hline
 Study&Modality&Valance&Arousal\\
 \hline
 \multicolumn{4}{c}{\textbf{MAHNOB-HCI}} \\
Ferdinando et al. \cite{17} & ECG  &42.55    &47.69\\
Yu et al. \cite{76} & HRV  & \textbf{46.86}    &44.02\\
\textbf{Our (Scalogram)} & ECG  &42.91 & \textbf{49.14}\\
 \hline
 \multicolumn{4}{c}{\textbf{DEAP}} \\
Patras et al. \cite{32} & EEG & 62.00 & 57.60 \\
Chung et al. \cite{74} & EEG & 70.90 & 70.10 \\
Campos et al. \cite{75} & EEG & 73.14 & 73.06 \\
Lee et al. \cite{36} & PPG & 75.30 & 76.20 \\
Elalamy et al. \cite{56} & PPG(spectrogram) & 69.60 & 69.30 \\
Elalamy et al. \cite{56} & PPG(Recurrence Plots) & 69.40 & 69.20 \\
\textbf{Our (Scalogram)} & PPG    &\textbf{76.51} & \textbf{77.02}\\
 \hline
\end{tabular}
\end{center}
\end{table}
%\vspace{-2.5cm}
\begin{table}
\caption{Performance comparison of our multimodal emotion recognition method with the state-of-the-art techniques.}
\label{table:4}
\begin{center}
%\begin{tabular}{l p{4.5cm} c c} 
\begin{tabular}{l p{2.4cm} c c} 
 %\hline
 %\multicolumn{4}{|c|}{Country List} \\
 \hline
 Study&Modality&Valance&Arousal\\
 \hline
 \multicolumn{4}{c}{\textbf{MAHNOB-HCI}} \\
Soleymani et al. \cite{31} & EEG and Eye Gaze  &\textbf{76.10}    &67.70\\
Koelstra et al.  \cite{15} & EEG and Face  &73.00    &68.50\\
\textbf{Our (UBVMT)} & ECG and Face  &50.01 & \textbf{83.84}\\
 \hline
 \multicolumn{4}{c}{\textbf{DEAP}} \\
Yin et al. \cite{53} & EEG, ECG, EOG, GSR, EMG, Skin temperature, Blood
volume, Respiration  & 76.17    &77.19\\
Shang et al. \cite{13} & EEG, EOG, EMG & 51.20 & 60.90 \\
Siddharth et al. \cite{7} & EEG, PPG, GSR & 71.87 & 73.05 \\
Siddharth et al. \cite{7} & EEG and Face & 73.94 & 74.13 \\
Elalamy et al. \cite{56} & PPG, EDA & 69.90 & 69.70 \\
\textbf{Our (UBVMT)} & PPG and Face    &\textbf{81.53} & \textbf{82.64}\\
 \hline
\end{tabular}
\end{center}
\end{table}

\begin{table}
\caption{F1 scores of our multimodal emotion recognition method.}
\label{table:5}
\begin{center}
\begin{tabular}{l p{2.4cm} c c} 
 %\hline
 %\multicolumn{4}{|c|}{Country List} \\
 \hline
 Study&Dataset&Valance&Arousal\\
 \hline

\textbf{UBVMT} & MAHNOB-HCI &0.501 & 0.846\\
 \hline
\textbf{UBVMT} & DEAP    &0.815 & 0.829\\
 \hline
\end{tabular}
\end{center}
\end{table}

\begin{figure}[t]
\centering
\includegraphics[width=8.5cm, height=6.0cm]{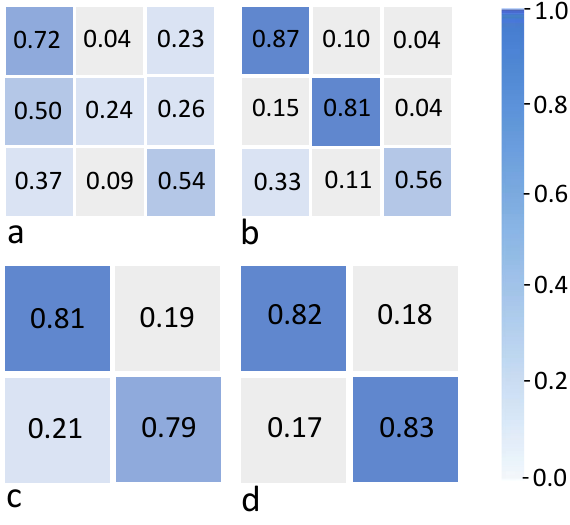}
\vspace{-0.4cm}
\caption{The confusion matrices for (a). MAHNOB-HCI valence emotion recognition, (b). MAHNOB-HCI Arousal emotion recognition, (c). DEAP valence emotion recognition, and (d). DEAP arousal emotion recognition.}
\label{fig:5}
%\vspace{-0.5cm}
\end{figure}

\subsection{Results and Analysis of Multi-modal Method}

The average of the 10-fold cross-validation accuracy of the UBVMT multimodal emotion recognition method is shown in Table \ref{table:4}. However, since our method provides a baseline for multimodal emotion recognition by fusing the ECG/PPG signal with the face information, there are no state-of-the-art methods with which we can compare our results. Nonetheless, as can be seen in Table \ref{table:4} that the proposed method outperforms the multimodal techniques fusing EEG information with other physiological signals, especially for the recognition of arousal. For the MAHNOB-HCI dataset, we compare our results with the EEG-based multimodal methods proposed by Soleymani et al. \cite{31} and Koelstra et al. \cite{15}. Here again, the proposed method outperforms both of these techniques by a large margin in the recognition of arousal by producing an average accuracy of $83.84\%$. In contrast, the method proposed by Soleymani et al. \cite{31} achieves the highest accuracy of $76.10\%$ for the classification of the valence class. We argue that the superior performance of our technique in recognizing the arousal class is because UBVMT is pre-trained by employing an rPPG signal, and it has been reported in past research that PPG signals are more effective in categorizing arousal \cite{29}. Therefore, given a large enough ECG dataset to pre-train the UBVMT network, we posit that we can improve the classification accuracy of our method for the valence class as well. For the DEAP dataset, we compare our results with the multimodal techniques fusing PPG information with other signals like EEG, EDA GSR, etc. As can be seen in Table \ref{table:4}, our method outperforms the state-of-the-art multimodal emotion classification techniques both in recognizing valence and arousal with average accuracies of $81.53\%$ and $82.64\%$, respectively. It is interesting to note that our bimodal method outperforms the multimodal technique proposed by Yin et al. \cite{53}, where the information from EEG, ECG, EOG, GSR, EMG, Skin temperature, blood volume, and respiration signals are fused to classify emotions. Similarly, comparing our technique with the bimodal method proposed by Siddharth et al. \cite{7} where EEG and face information are fused for emotion analysis, Table \ref{table:4} shows that the fusion of PPG and face information using UBVMT is more effective in recognizing emotions. Our PPG-face method is also more accurate in classifying emotions compared to the fusion of EEG, EOG, and EMG signals \cite{13}. In \cite{56}, similar to our approach, Elalamy et al. transformed the 1D PPG and EDA signals into 2D representations and fuse them to categorize emotions using deep networks. Table \ref{table:4} shows that our scalogram-based PPG representation, when fused with the face data, produces higher recognition accuracy by employing our proposed technique. The confusion matrices and the F1 scores of the multimodal emotion recognition results are shown in Figure \ref{fig:5} and Table \ref{table:5}, respectively. 

\section{Conclusion}

In this paper, a novel approach called Unified Biosensor-Vision Multi-modal Transformer-based (UBVMT) method is presented to classify emotions in an arousal-valence space. The proposed technique combines a 2D representation of an ECG/PPG signal with facial information for emotion recognition. Initially, we investigated and compared the unimodal emotion recognition performance using three image-based representations of the ECG/PPG signal. Then, the Unified Biosensor-Vision Multi-modal Transformer-based network is used for emotion recognition by combining the 2D image-based representation of the ECG/PPG signal and facial information. Our unified transformer model comprises homogeneous transformer blocks that take the 2D representation of the ECG/PPG signal and associated face frame as input for emotion representation learning with minimal modality-specific design. The UBVMT model is trained using a reconstruction loss involving masked patches of video frames and 2D images of ECG/PPG signals, along with contrastive modeling to align face and ECG/PPG data. Extensive experiments on the MAHNOB-HCI and DEAP datasets demonstrate that our Unified Biosensor-Vision Multi-modal Transformer-based model achieves comparable results to state-of-the-art techniques.

% if have a single appendix:
%\appendix[Proof of the Zonklar Equations]
% or
%\appendix  % for no appendix heading
% do not use \section anymore after \appendix, only \section*
% is possibly needed

% use appendices with more than one appendix
% then use \section to start each appendix
% you must declare a \section before using any
% \subsection or using \label (\appendices by itself
% starts a section numbered zero.)
%

% use section* for acknowledgment
\ifCLASSOPTIONcompsoc
  % The Computer Society usually uses the plural form
  \section*{Acknowledgments}
\else
  % regular IEEE prefers the singular form
  \section*{Acknowledgment}
\fi

The authors would like to thank Mustansar Fiaz, Sachin Shah, Robinson Vasquez, Lisa Dieker, Shaunn Smith. Rebecca Hines, Ilene Wilkins, Kate Ingraham, Caitlyn Bukaty, Karyn Scott, Eric Imperiale, Wilbert Padilla, Maria Demesa for the discussions and suggestions throughout this research. This research was supported in part by grants from the National Science Foundation Grants 2114808, and from the U.S. Department of Education Grants H327S210005, H327S200009. The lead author (KA) was also supported in part by the University of Central Florida's Preeminent Postdoctoral Program (P3). Any opinions, findings, conclusions, or recommendations expressed in this material are those of the authors and do not necessarily reflect the views of the sponsors.

% Can use something like this to put references on a page
% by themselves when using endfloat and the captionsoff option.
\ifCLASSOPTIONcaptionsoff
  \newpage
\fi

\vspace*{-3.0cm}

% biography section
% 
% If you have an EPS/PDF photo (graphicx package needed) extra braces are
% needed around the contents of the optional argument to biography to prevent
% the LaTeX parser from getting confused when it sees the complicated
% \includegraphics command within an optional argument. (You could create
% your own custom macro containing the \includegraphics command to make things
% simpler here.)
\begin{IEEEbiography}[{\includegraphics[width=1in,height=1.25in,clip,keepaspectratio]{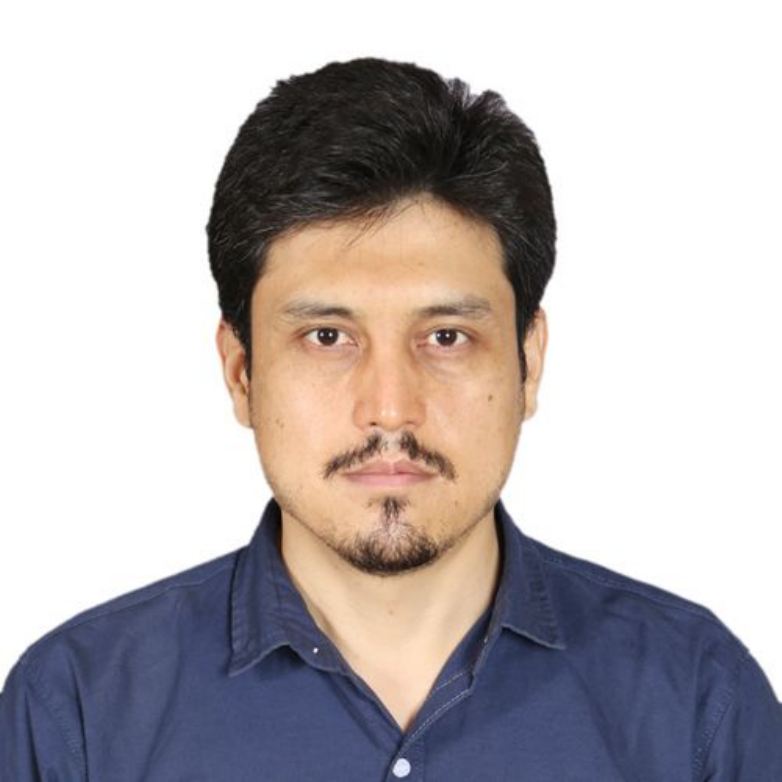}}]{Kamran Ali}
Kamran Ali is a postdoctoral research associate in the computer science department at the University of Central Florida, Orlando, FL, 32816, USA. His research interests include computer vision and machine learning. He is the corresponding author of this article. Contact him at kamran.ali@ucf.edu.
% or if you just want to reserve a space for a photo:
\end{IEEEbiography}
%\vspace{1pt} % Adjust the value as needed
\vspace*{-3cm}
\begin{IEEEbiography}
[{\includegraphics[width=1in,height=1.25in,clip,keepaspectratio]{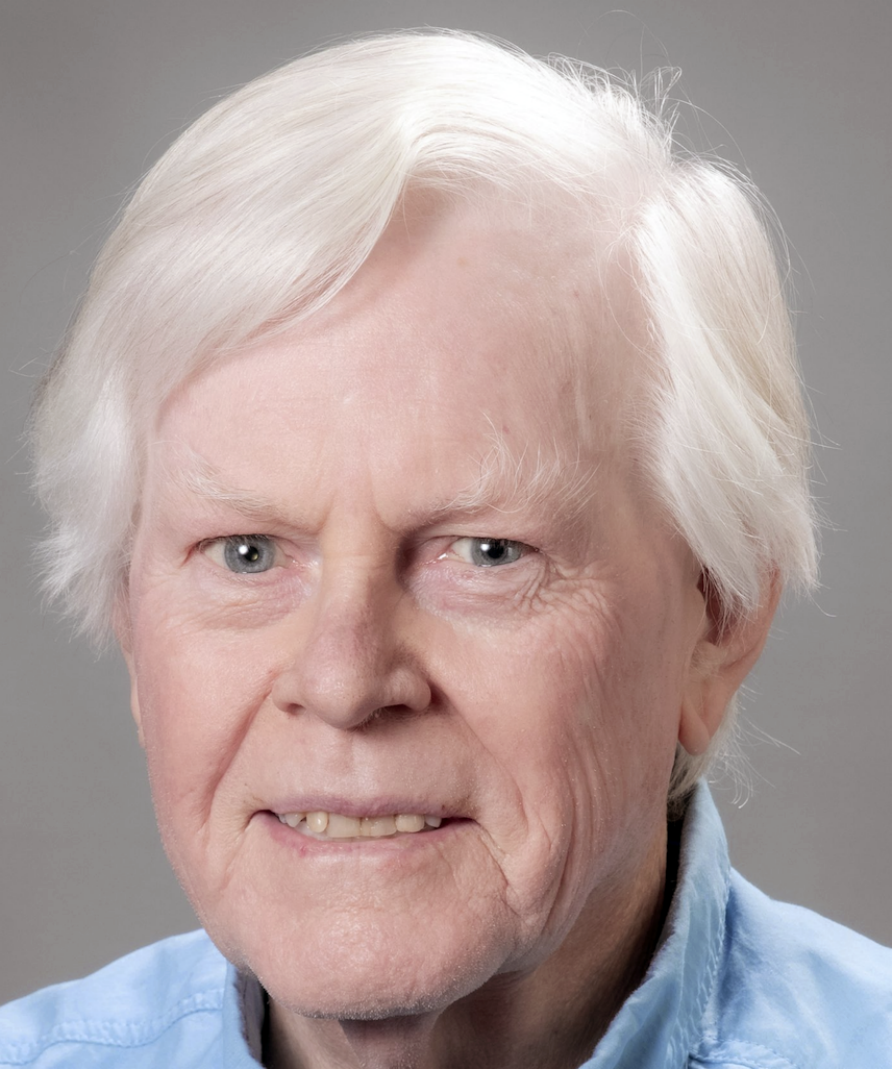}}]{Charles E. Hughes}
Charles E. Hughes is a pegasus professor of Computer Science at the University of Central Florida, Orlando, FL, 32816, USA. His research interests include virtual learning environments, computer graphics, machine learning, and visual programming systems. Contact him at charles.hughes@ucf.edu.

\end{IEEEbiography}
%\begin{IEEEbiography}{Michael Shell}
%Biography text here.
%\end{IEEEbiography}

% insert where needed to balance the two columns on the last page with
% biographies
%\newpage

% You can push biographies down or up by placing
% a \vfill before or after them. The appropriate
% use of \vfill depends on what kind of text is
% on the last page and whether or not the columns
% are being equalized.

%\vfill

% Can be used to pull up biographies so that the bottom of the last one
% is flush with the other column.
%\enlargethispage{-5in}

% that's all folks
\end{document}